\documentclass[a4paper, 10pt, conference]{ieeeconf}      

\IEEEoverridecommandlockouts                              

\usepackage{array,multirow,graphicx} 
\usepackage{cite}
\usepackage{amsmath}
\usepackage{amssymb}

\usepackage{bm}
\usepackage{bbm}
\usepackage{siunitx}
\usepackage{tikz}
\usepackage{pgfplots}					
\usepackage{pgfplotstable}
\usepgfplotslibrary{statistics}		
\pgfplotsset{width=7cm,compat=newest, every tick label/.append style={font=\tiny}}	
\usepackage{filecontents}
\usepackage{todonotes}
\usepackage{pdfpages}
\usepackage{pgf}
\usetikzlibrary{calc,matrix}
\newcommand\myeq{\mkern1.5mu{=}\mkern1.5mu}
\usetikzlibrary{external}

\title{\LARGE \bf
A Multi-Stage Clustering Framework for Automotive Radar Data
}

\author{Nicolas Scheiner$^{1}$, Nils Appenrodt$^{1}$, J\"urgen Dickmann$^{1}$, and Bernhard Sick$^{2}$
\thanks{$^{1}$Daimler AG, Wilhelm-Runge-Str. 11, 89081 Ulm, Germany
        {\tt\small nicolas.scheiner@daimler.com}}
\thanks{$^{2}$Intelligent Embedded Systems, University of Kassel, Wilhelmsh\"oher Allee 73, 34121 Kassel, Germany
        {\tt\small bsick@uni-kassel.de}}%
}

\begin{document}

\maketitle
\thispagestyle{plain}
\pagestyle{plain}

\begin{abstract}
Radar sensors provide a unique method for executing environmental perception tasks towards autonomous driving.
Especially their capability to perform well in adverse weather conditions often makes them superior to other sensors such as cameras or lidar.
Nevertheless, the high sparsity and low dimensionality of the commonly used detection data level is a major challenge for subsequent signal processing.
Therefore, the data points are often merged in order to form larger entities from which more information can be gathered.
The \emph{merging} process is often implemented in form of a clustering algorithm.
This article describes a novel approach for first filtering out static background data before applying a two-stage clustering approach.
The two-stage clustering follows the same paradigm as the idea for data association itself:
First, clustering what is ought to belong together in a low dimensional parameter space, then, extracting additional features from the newly created clusters in order to perform a final clustering step.
Parameters are optimized for filtering and both clustering steps.
All techniques are assessed both individually and as a whole in order to demonstrate their effectiveness.
Final results indicate clear benefits of the first two methods and also the cluster merging process under specific circumstances.
\end{abstract}

\section{Introduction} \label{seq:intro}
\begin{figure}[t!]
	\centering	
	\includegraphics[width=1.\columnwidth]{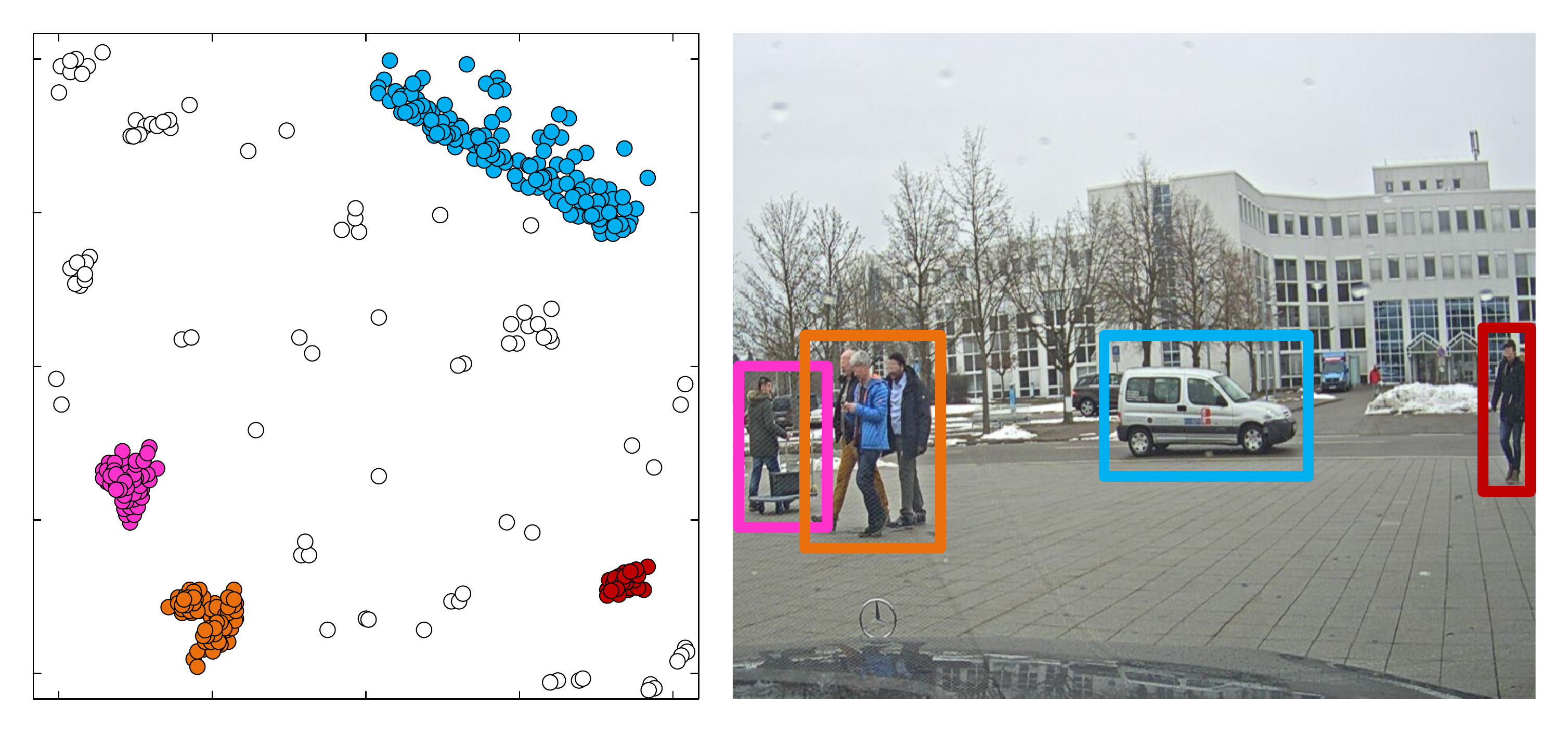}
	\caption{Radar data points and reference image of typical clustering task. The objects of interest are indicated by the same color in the two images.}	
	\label{fig:abs_problem}
\end{figure}

\begin{figure}[b!]
	\centering	
	\includegraphics[width=1.\columnwidth]{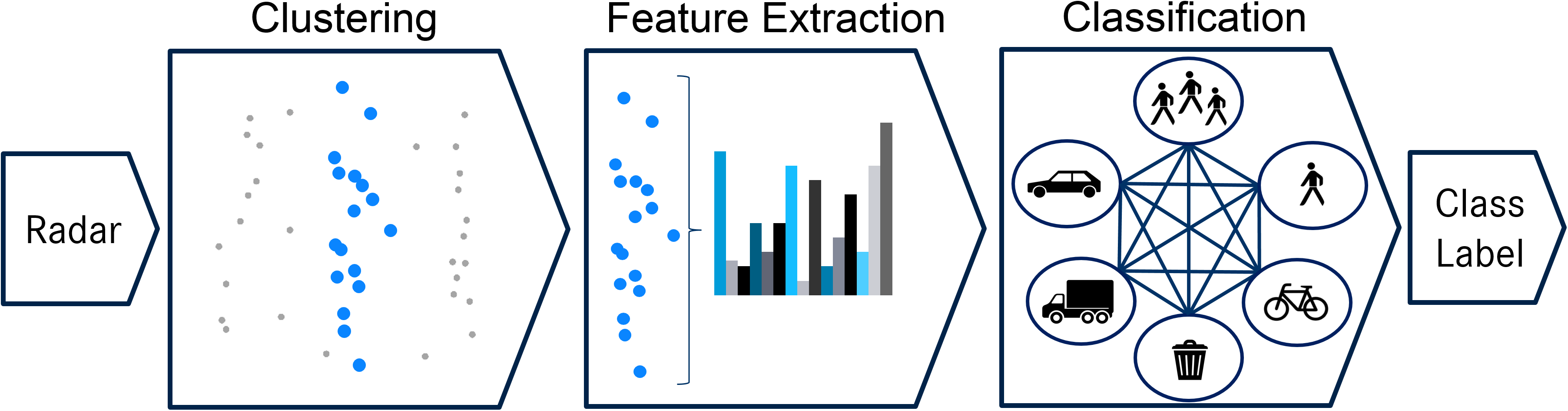}
	\caption{Conventional radar data classification process as used in \cite{Scheiner2018}: data is first transformed to a common coordinate system where clusters are formed. Then, feature extraction and classification are performed on each clusters, outlining the importance of the actual clustering process.}	
	\label{fig:clf_chain}
\end{figure}

Radar sensing is an integral part of current perception concepts for autonomously driving vehicles.
This is justified by a radar's ability to directly obtain a precise radial (Doppler) velocity from all observed objects within a single measurement.
It is the only automotive sensor to deliver this information in a single shot fashion and, therefore, is indispensable for adaptive cruise control systems \cite{winner2016}.
Moreover, state-of-the-art automotive radar sensors typically operate at a frequency range of $76-$\SI{81}{GHz} which makes them more robust to adverse weather conditions such as fog, snow, or heavy rain.
The drawback of radar is its low angular resolution in comparison to other sensors.
This leads to a relatively sparse data representation, especially for remote objects.
The final goal of most perception tasks is semantic instance segmentation, i.e., object instances in the radar data have to be identified and classified as depicted for moving road users in Fig. \ref{fig:abs_problem}.
In order to decide which data points make up an object instance, static radar object detection is able to accumulate data of several measurement cycles to build grid maps as a compensation for the data sparsity \cite{Lombacher2017}.
A common approach for identifying moving road users is to utilize clustering algorithms for data grouping.
Once clusters are formed, the additional information that can be extracted from aggregated data points is beneficial for classification algorithms as shown, e.g., in \cite{Schubert2015, Schumann2017, Scheiner2018, Prophet2018, Scheiner2019iv} and summarized in Fig. \ref{fig:clf_chain}.
The presented techniques are usually based on radar sensors delivering detection points.
Those detections are the result of a constant false alarm rate (CFAR) detector \cite{richards2005} which returns only those reflection points that exceed a self-adaptive reflection amplitude threshold.
While CFAR filtering is certainly a powerful method for reducing the amount of data to a manageable level, it also returns many reflections which do not correspond to any object of interest.
This article is focused solely on moving road users, hence it is reasonable to filter out detections that correspond to static objects prior to clustering.
To this end, this article proposes a filtering concept that relies on detections' Doppler information and spatial density for reducing the number of data points to be clustered.
Furthermore, a two-stage clustering algorithm is elaborated in detail, including a discussion of several implicit decisions that are often neglected in literature.
The two-stage clustering consists of an ordinary clustering algorithm that is improved in a first stage.
\begin{figure*}[t!]
	\centering	
	\includegraphics[width=1.\linewidth]{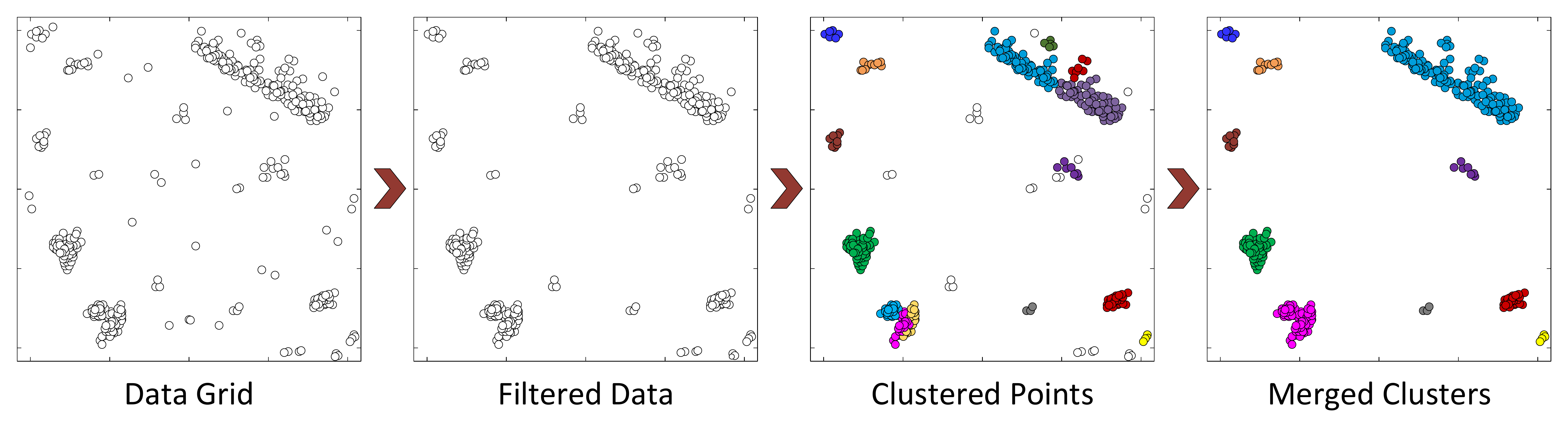}
	\caption{Radar data points and reference image of typical clustering task. In both images the objects of interest are indicated by matching colors.}	
	\label{fig:overview}
\end{figure*}
In the second stage, preliminary clusters from stage 1 are merged using domain knowledge to extract additional features from the combined information contained in the pre-clustered data points.
The complete processing chain is illustrated in Fig. \ref{fig:overview}.
After parameter optimization in both steps on a an extensive automotive data set clear benefits of detection filtering and the proposed improvements for the first clustering step can be determined.
The second cluster merging step is advantageous only under certain conditions which will be discussed as well.

The article is organized as follows: In Section \ref{sec:rel_work} basic clustering concepts and related work are discussed.
Section \ref{sec:methods} explains how data filtering and clustering algorithms are designed and how parameter optimization is implemented.
Experiments for all individual processing steps are described and evaluated in Section \ref{sec:results}.
Section \ref{sec:conclusion} concludes the topic and gives prospects for future work.

\section{Related Work}\label{sec:rel_work}
\begin{figure}[b!]
\centerline{\includegraphics{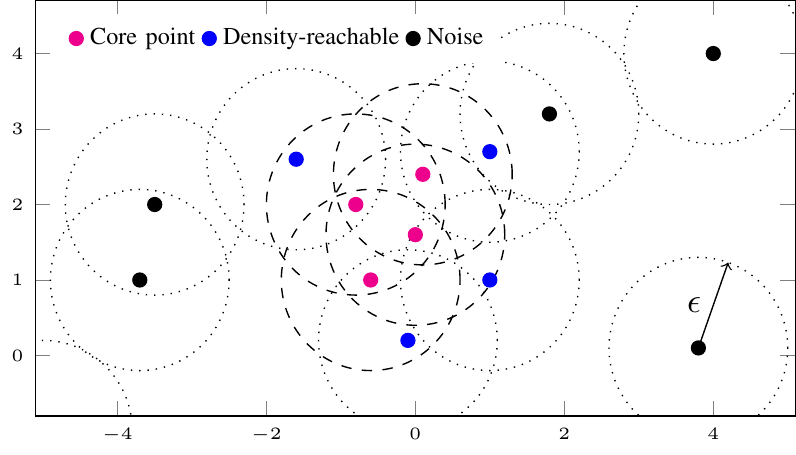}}
\caption{DBSCAN example plot in two-dimensional Euclidean space. $\epsilon$-regions around each point are indicated with circles, $N_{\text{min}}=2$. The attributed point class, i.e., core, border, or noise is color-coded.}
\label{fig:dbscan}
\end{figure}

Radar data association is a challenging task.
One reason for this is that radar reflection points are the product of a random process which makes it hard to define absolute distance or density criteria.
When using a clustering algorithm to segment the radar scene, the palette of techniques is wide.
The algorithm needs to fulfill certain speed criteria, it must not be limited in the maximum number of resulting cluster instances, and it has to cope with varying densities between associated detections and clutter in close distance.
The DBSCAN algorithm \cite{Ester1996} is a fast and elegant method which has widely been accepted among data scientists working with radar data.
A short example of the algorithm's mode of operation is given in Fig. \ref{fig:dbscan}.
Essentially, for each point in a search space the number of close-by neighbors is determined using an appropriate distance measure.
The neighborhood is defined by a multidimensional maximum threshold value $\epsilon$.
A cluster is created if a point has a minimum number of neighbors $N_{\text{min}}$.
The corresponding data sample is called a \emph{core point}.
All points that fall within an $\epsilon$-region of a core point are added to the cluster.
If a sample is added to a cluster which is not a core point itself, it is called \emph{density-reachable}.
The remainder of samples is classified as \emph{noise}.

Many authors have made modifications to the DBSCAN algorithm for utilizing it to cluster radar data.
In \cite{Kellner2012} it is shown how the radar's propagation of radiation and the consequential change in sampling density can be incorporated in the DBSCAN algorithm.
Another approach for radar clustering is presented in \cite{Schubert2015} where the problem is simplified by projecting the data in an $x/y/v_r$ grid.
In that article, data below an absolute threshold of \SI{0.1}{\meter} is filtered out.
This concept is further refined in \cite{Schumann2018b} where the time dimension is added to the data grid and parameter optimization on different distance-velocity regions is used to generate several sets of clustering coefficients.
Furthermore, instead of filtering out detections with low radial velocity, the authors propose simply not allowing them as core points.
They also investigate a hierarchical DBSCAN variant \cite{McInnes2017}, but find its performance inferior to the original algorithm.
In contrast to previous authors, in \cite{Prophet2018} it is argued that the variations in Doppler velocity are often too large for use as a cluster criterion.
Another approach to automotive radar clustering was presented in \cite{Wagner2015}, where the power levels of signals are utilized to calculate more precise distances in the spatial and Doppler dimension.
As that approach requires access to a deeper data level than detections, this method cannot be used here. 

In this article, the DBSCAN algorithm is evaluated using different distance metrics based on an $x/y/v_r/t$ grid.
Epsilon values for the individual parameter dimensions are optimized with the assumption that a single fixed set should suffice.
Equally, real world road users' extents are expected to remain in similar ranges of values.
To enable static epsilon values, beam dilation is counteracted using a range-dependent value for $N_{\text{min}}$.
Splitting the clustering process into two separate DBSCAN steps, should relax the requirements for the clustering algorithm to the point where additional information can be extracted from the first clustering results.
The extra information is then used to merge clusters that would have required extensive parameterizations in single step clustering.
Furthermore, important preprocessing steps for merging data from several sensor scans and filtering out unnecessary data are presented.

\section{Methods}\label{sec:methods}
As discussed above, the proposed framework consists of three components that are discussed individually in this section:
\begin{enumerate}
\item \textbf{Detection filtering}
\item \textbf{Point clustering}
\item \textbf{Cluster merging}
\end{enumerate}
Additionally, an extra subsection covers parameter optimization of the clustering methods and the definition of a \emph{target} -- or \emph{score} -- function.

\subsection{Detection Filtering}
In the initial step, the detections are filtered.
This process serves two purposes:
First, subsequent processing steps are accelerated when the number of input points is reduced.
This can lead to huge time savings in case a cost-effective filtering routine is used.
The second advantage of filtering is that undesirable detections close to objects of interest can harm the performance of a clustering algorithm. 
Cluster parameters can be selected more freely by removing many of these points.

The main idea behind the proposed filtering method is the following:
Road users in motion have a high probability of retaining several reflections after the CFAR detector due to their velocity which distinguishes them from stationary objects.
Lateral movements towards a single sensor can result in minuscule Doppler velocities.
The test vehicle uses multiple distributed radar sensors which retain the radial components of these movements.
Hence, there are two properties that can serve as decision criterion for a filtering algorithm:
Doppler velocity and detection density.
It should be noted that neither criterion is sufficient for making a hard choice about the membership of a detection to an object of interest.
Therefore, radial velocity $v_r$ and amount of neighbor points $N(d_{xy},\Delta t)$ in a spatial and a temporal distance $d_{xy}$ and $\Delta t$ around each point are determined and used as a combined assessment factor.
The fewer the number of neighbors, the higher the radial velocity of a point needs to be.
While $\Delta t$ is set to a fixed value of \SI{0.25}{\second} corresponding to the time windows in following steps, $d_{xy}$ can be tuned to maximize the filter's performance.
A detection is then filtered out when the following criterion holds:
\begin{equation}\label{eq:filter_metric}
\resizebox{.89\linewidth}{!}{%
$\begin{split}
N(d_{xy}) < 1 \:\vee\: (v_r < \mathbf{\eta}_{\mathbf{v}_{\mathbf{r},i}} \:\wedge\: N(d) < \mathbf{N}_i) \quad \forall i\in\{1,\dots ,4\}, \\
\text{with } \eta_{v_{r},1} = 5\cdot \eta_{v_{r},2} = 10\cdot \eta_{v_{r},3} = 50\cdot \eta_{v_{r},4}, \\
\text{ and } \mathbf{N} = (2, 3, 4, 10).
\end{split}$
}
\end{equation}
The threshold ratios for $\mathbf{\eta_{v_{r}}}$ and $\mathbf{N}$ have been selected empirically in order to reduce the number of tuning parameters to only two.
During optimization the maxim is to eliminate as many points as possible without removing relevant ones, i.e., detections belonging to objects of interest.
Given the fixed assessment metric in Eq. \ref{eq:filter_metric}, values for the threshold $\eta_{v_{r},1}$ and $d_{xy}$ can be estimated by allowing a fixed amount of falsely removed detections per cluster.
For the preservation of a high amount of detections on moving road users, parameters are adjusted to result in a maximum of \SI{20}{\%} wrongly removed detections per cluster at any given time frame.
This ensures that at any given time, subsequent processing steps have a sufficient number of detections to identify road users.
In order to determine the distance between radar detection points from different time steps all points require a shared reference.
A common approach for fusing sensory data from different automotive sensors is to transform all data to a fixed coordinate system with origin at the rotational center -- usually the middle of rear axle -- of the vehicle.
The corresponding coordinate system is referred to as the \emph{car coordinate system} (CCS).
To transform the sensor data to the CCS, the relative installation position and orientation relative to the CCS suffice to determine a fixed transformation matrix.
When using temporal data the transformation is more complicated unless the assumption can be made that the ego-vehicle did not move substantially far during a time frame.
If this is not the case, a common coordinate system requires incorporating the ego-vehicle's offset in spatial position and orientation during all time steps in the regarded frame.
In theory, the resulting frame coordinate system (FCS) is superior to simple CCS processing.
To also ensure practical relevance, filtering and the first clustering step will both be examined in each coordinate system.

\subsection{Point Clustering} \label{subsec:step1}
The actual detection clustering process follows.
In this step, the radar points remaining after the filtering process are associated by utilizing a slightly modified version of a DBSCAN algorithm.
The main differences to conventional DBSCAN is an adaptive number of minimum points $N_{\text{min}}(r)$ required to form a cluster core point, where $r$ is the range, i.e., distance between detection and sensor.
This adjustment is based on the fact that remote objects have a smaller maximum number of possible reflections due to range-independent angular resolution as discussed in \cite{Kellner2012}.
Since the physical extents of road users do not change, this article utilizes only the minimum point property to account for remote objects:
\begin{equation} \label{eq:min_pts}
\resizebox{.89\linewidth}{!}{%
$
N_{\text{min}}(r) = N_{\text{min},\SI{50}{\meter}} \cdot \left(1 + \alpha_r \cdot \left(\frac{\text{clip}(r,\SI{25}{\meter},\SI{125}{\meter})}{\SI{50}{\meter}}-1\right)\right).
$}
\end{equation}
Eq. \ref{eq:min_pts} has two tuning parameters: $N_{\text{min},\SI{50}{\meter}}$ and $\alpha_r$ which represent a minimum point baseline at $\SI{50}{\meter}$ and the slope of the reciprocal relation.
To avoid very low or high numbers for $N_{\text{min}}(r)$ the range $r$ is clipped to values between \SIlist{25;125}{\meter}.

Furthermore, only detections that exceed a certain radial velocity threshold $v_r > v_{r,\text{min}}$ can become \emph{core points} in accordance with \cite{Schumann2018b}.

Another major question to addressed in this step is the choice of distance metric or neighborhood criterion.
The four considered variables for point clustering are: Doppler velocity, time, x-range, and y-range.
As amplitudes often have very high variations even on a single object they are neglected.
For this article, three neighborhood criteria have been examined.
They all treat the time $t$ as an independent variable, i.e., $\Delta t$ is always required to be smaller or equal than its corresponding threshold $\epsilon_t$.
In essence, this has the same effect as a sliding window which is used for real-time processing.
In the same manner, the first neighborhood criterion combines the differences $\Delta\cdot$ of all four variables:
\begin{align} \label{eq:step1_1}
\Delta x < \epsilon_x \:\wedge\: \Delta y < \epsilon_y \:\wedge\: \Delta v_r < \epsilon_{v_r} \:\wedge\: \Delta t < \epsilon_t.
\end{align}
According to \cite{Schumann2018b}, the spatial threshold values are set to the same value $\epsilon_x = \epsilon_y = \epsilon_{xy}$ for more stability for rotated objects.
This method serves as baseline for further variants.
The second method aims to achieve full rotational invariancy by utilizing the Euclidean distance of the spatial components:
\begin{align} \label{eq:step1_2}
\sqrt{\Delta x^2 + \Delta y^2} < \epsilon_{xy} \:\wedge\: \Delta v_r < \epsilon_{v_r} \:\wedge\: \Delta t < \epsilon_t.
\end{align}
In principle, it is reasonable to allow higher velocity offsets for close detections than for remote ones and vice versa.
Therefore, the third criterion combines both spatial components and the radial velocity in a single Euclidean distance:
\begin{align} \label{eq:step1_3}
\sqrt{\Delta x^2 + \Delta y^2 + \frac{1}{\epsilon^{'2}_{v_r}}\cdot\Delta v_r^2} < \epsilon_{xyv_r} \:\wedge\: \Delta t < \epsilon_t.
\end{align}
In this case $\epsilon'_{v_r}$ and $\epsilon_{xyv_r}$ have the same scaling effect as $\epsilon_{v_r}$ and $\epsilon_{xy}$ in Eqs. \ref{eq:step1_1} and \ref{eq:step1_2}.
However, they do not represent absolute maximum velocity or spatial distance thresholds anymore.
The scaling of ${v_r}$ allows for better tuning capabilities than, e.g., normalizing all values to the same range.
The time $t$ is not included in Euclidean distance due to real-time processing constraints as mentioned above.
Also, $\epsilon_{t}$ has a high influence on the number of points accumulated from the radar, i.e., tuning $N_{\text{min}}$ would require taking into account sensor cycles as well.

\subsection{Cluster Merging}
In the second clustering step, only those points are considered that have been assigned a cluster label in the previous step.
The key idea for this step is that a cluster contains more information than its individual detections on their own.
Therefore, cluster merging is based on different DBSCAN parameters than the first clustering step.
Specifically, the goal is to calculate information about the cluster's moving speed and orientation.
To this end, two concepts are shortly discussed and investigated:
\paragraph{Object velocity estimation}
By definition, radial velocities always point radially towards the measuring radar sensor.
The length and orientation of the real velocity vector are unknown.
In order to estimate the real object velocity, multiple detections belonging to the same object must be available.
For rigid objects, theoretically two detections that either occur at different azimuth angles or are captured by two different sensors suffice.
In practice however, a more stable solution can be found using additional detections.
In \cite{Kellner2013}, a system for real object velocity estimation was proposed.
The system utilizes an outlier filtering algorithm to remove faulty detections.
Then, an optimization problem is solved that yields a single velocity vector for the object of interest.
This concept is applied to the clusters estimated in the previous subsection.
Therefore, the neighborhood criterion of a secondary clustering step is formulated as:
\begin{equation} \label{eq:real_vel_metric}
d_\text{min}<\epsilon_d \:\wedge\: \Delta\phi < \epsilon_{\phi} \:\wedge\: \Delta v < \epsilon_v \:\wedge\: \Delta t < \epsilon_{t2},
\end{equation}
where $d_\text{min}$ is the Euclidean distance between the two closest detection members of corresponding clusters during the observed time frame which is defined by $\epsilon_{t2}$.
$\Delta\phi$ and $\Delta v$ denote the differences in velocity orientation and magnitude, respectively.
\paragraph{Spatial cluster continuation}
Instead of relying on the distribution of radial velocities, additional information can be obtained by regarding the temporal progression of spatial detection distributions.
To this end, spatial cluster centers are calculated for every given time step based on the average $x/y$ coordinates.
Then, a smooth trajectory is fitted by first applying a moving average filter to the temporal distribution of the cluster centers before using cubic -- or linear when cubic is not possible -- spline interpolation.
The splines are used for resampling the trajectory at a finer rate.
The gradient of the resampled trajectory then serves as a velocity approximation which is used to predict future cluster centers based on a continuity assumption.
The neighborhood criterion for this method is defined as:
\begin{equation}
d_\text{pred,min} < \epsilon_{d} \:\wedge\: \Delta v < \epsilon_{v} \:\wedge\: \Delta t < \epsilon_{t2}.
\end{equation}
$d_\text{pred}$ is the Euclidean distance of the two compared clusters' predicted centers.
This value is calculated at three fixed times at the beginning, in the middle, and at the end of the time frame $\epsilon_{t2}$.
Only the minimum distance of those three samples is used.
In the same manner, the predicted velocity offset for $\Delta v$ is estimated.

\subsection{Parameter Optimization}
For all metric variants in clustering steps 1 and 2, the threshold and scaling parameters are optimized using Bayesian Optimization \cite{Mockus1974}.
Bayesian Optimization is a derivative-free optimization strategy that aims to find an optimal parameter set which maximizes the output of a surrogate function.
The surrogate is built to resemble the objective function which is unknown and expensive to sample.
Optimization is split into two parts:
In the \emph{exploration} phase the parameter space is tested for promising areas.
Then, those areas are examined more closely during \emph{exploitation}.
In order to optimize the clustering process, a score function is required for result rating.

The score function is required to reward cluster formations that are similar to the labeled ground truth.
Specifically, it is important to form an explicit cluster for every road user instance which should contain as many related points as possible.
It is, however, also necessary to stop clustering at an object's end, i.e., not to merge clusters from different road users or to add detections from surrounding background to the clusters of interest.
The majority of the data points in a radar scene will be background detections.
For those detections it is beneficial not to form clusters at all.
As subsequent classification stages learn to distinguish road users from nonrelevant cluster formations it is not crucial to prevent these clusters at the cost of other -- more relevant -- ones.
For this article the \emph{V-measure} \cite{rosenberg2007} is chosen as a suitable score function as it combines several of the named requirements.
The V-measure is a combined score based on two intuitive clustering criteria, \emph{homogeneity} and \emph{completeness}.
Homogeneity is maximal when a predicted cluster only contains points from a single ground truth cluster. 
Contrary to that, completeness aims to assign all points from a single ground truth cluster into a single cluster prediction.
Both are based on the conditional entropy of predicted clusters $K$ given ground truth clusters $C$ and vice versa:
\begin{align} \label{eq:homocomp}
\text{Homogeneity} &= 1 - \frac{H(C|K)}{H(C)}, \\
\text{Completeness} &= 1 - \frac{H(K|C)}{H(K)}, \label{eq:comp}
\end{align}
\begin{equation*}
\begin{split}
\text{with} \quad H(A|B) &= - \sum_{a\in A} \sum_{b\in B} \frac{n_{a,b}}{n}\cdot \log \frac{n_{a,b}}{n_b}, \\
\text{and} \quad H(A) &= - \sum_{a\in A} \frac{n_{a}}{n}\cdot \log \frac{n_{a}}{n}.
\end{split}
\end{equation*}
$H(A)$ and $H(A|B)$ represent the entropy and the conditional entropy, respectively.
$A$ and $B$ can be substituted for $C$ and $K$ in the required order to match Eqs. \ref{eq:homocomp} and \ref{eq:comp} with $n$ being the total number of samples, $n_a$ and $n_{a,b}$ being the number of samples belonging to $a$, or $a$ and $b$ at the same time.
The V-measure $V_1$ is the harmonic mean of homogeneity and completeness:
\begin{equation}\label{eq:vmeas}
V_1 = 2\cdot \frac{\text{Homogeneity}\cdot\text{Completeness}}{\text{Homogeneity}+\text{Completeness}}.
\end{equation}
In order to suppress the penalization for the creation of background clusters, the calculation of the completeness score assumes perfect matching of the detections that belong to a labeled object in the ground truth.
With this adaptation the score's objective is sufficiently similar to the requirements for automotive radar clustering.
Compared to other variants presented in literature, it has the advantage of better comparability within different scoring results.
Especially in cases where one or two splits within a cluster are preferred over massively overestimated cluster boundaries, the V-measure proves to be beneficial.
For more details on the score, refer to the original publication.

Bayesian Optimization is performed for each step separately with a maximum of 100 iterations (30 for exploration and 70 for exploitation) per experiment.
Early experiments showed, that the optimizers had trouble finding suitable parameters for both steps at the same time.
This problem is further enhanced by the increased computational complexity that occurs because intermediate results are hardly reusable for a combined 2-step optimization.

\section{Experiments \& Results}\label{sec:results}
In this section, the individual methods are evaluated and optimized parameter settings are reported.
All experiments are based on a real world data set including roughly a million detection points on over 1000 road users in motion.
Sensor specifications can be found in \cite{Scheiner2018} or \cite{Scheiner2019iv}.
The data set is split into two disjoint parts of almost equal size, one for parameter estimation (training set) and one for evaluation (test set).
The reported scores are based on the test set.
For the two clustering steps final scores are reported as $V_1$ scores.

\subsection{Filtering Results}
For the filtering parameter estimation, a full enumeration on all threshold combinations within reasonable ranges is executed.
For $\eta_{v_{r},1}$ the search space is spanned from \SI{0.05}{\meter\per\second} to \SI{0.35}{\meter\per\second} and for $d_{xy}$ from \SIrange{0.8}{2.0}{\meter}.
The main goal of filtering is to eliminate as many irrelevant detections as possible.
However, it is more important not to harm the performance of the following processing steps due to filtering.
Therefore, Tab. \ref{tab:filter_errors} depicts the number of times where a real world object did not retain at least \SI{75}{\%} of its detections during a whole time frame of \SI{150}{\milli\second}.
The time frame is chosen according to the feature extraction time frame in \cite{Scheiner2018}.
Objects that are not present in the data for at least \SI{150}{\milli\second} are neglected here.
These objects are most likely at the edge of the field of view and are on the verge of entering or exiting.
While this criterion results in some faulty removals, it is better suited to ensuring that an object can potentially be identified than, e.g., a fixed threshold of \SI{1}{\%} which is based on the whole object sequence.
As can be seen from Tab. \ref{tab:filter_errors}, several parameterizations lead to minimal errors.
The percentages of some promising settings are listed in Tab. \ref{tab:filter_scores}.
The chosen setting of \SI{0.1}{\meter\per\second} and \SI{1.4}{\meter} velocity and distance thresholds does not have the best overall removal score (\SI{29.10}{\%} opposed to \SI{31.47}{\%}).
It turns out, however, to be a good compromise between both thresholds and is, hence, assumed to be more stable for scenes with small and very slow moving road users.
Tab. \ref{tab:filter_scores} also indicates that using a ego-motion compensated frame coordinate system (FCS) results in better removal rates than simple CCS processing.
Moreover, allowing for some errors during filtering can obviously lead to even better background suppression.
However, for further evaluation a more conservative setting is preferred.

\begin{table}[tb]
\caption{Filtering result error matrix. For each parameter combination the number of optimization criterion violations is indicated. An asterisk indicates the chosen setting.}
\label{tab:filter_errors}
\centerline{\includegraphics{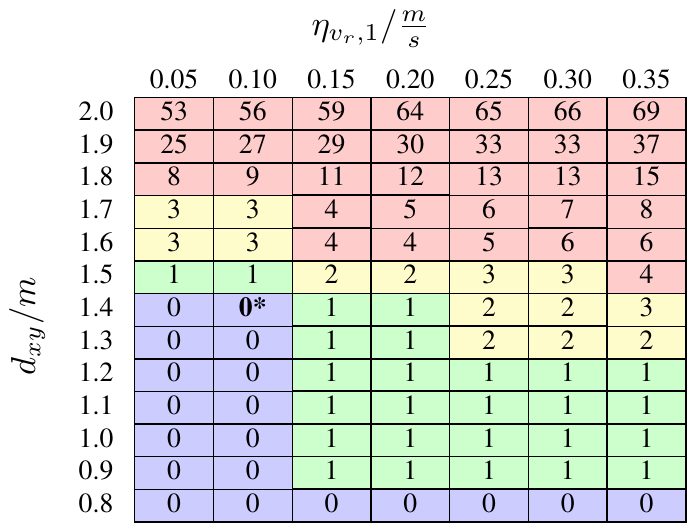}}
\end{table}

\begin{table}[tb]
\renewcommand{\arraystretch}{1.4}
	\caption{Detection filter settings for coordinate system (CS.), spatial, and neighbor thresholds along with the resulting background suppression rates and number false removals.}
	\label{tab:filter_scores}
	\centering
	\begin{tabular}{lll|ll}
		\hline
 		\textbf{CS.} & $\mathbf{\eta_{v_{r},1}}$ & $\mathbf{d_{xy}}$ & \textbf{Removal Rate} & \textbf{Errors} \\
 		\hline
		CCS & \SI{0.10}{\meter\per\second} & \SI{1.4}{\meter} &  \SI{25.10}{\%} & 0 \\
		\textbf{FCS} & \textbf{\SI{0.10}{\meter\per\second}} & \textbf{\SI{1.4}{\meter}} &  \textbf{\SI{29.10}{\%}} & \textbf{0} \\
		FCS & \SI{0.35}{\meter\per\second} & \SI{0.8}{\meter} &  \SI{31.47}{\%} & 0 \\
		FCS & \SI{0.20}{\meter\per\second} & \SI{1.4}{\meter} &  \SI{35.45}{\%} & 1 \\
		FCS & \SI{0.35}{\meter\per\second} & \SI{1.2}{\meter} &  \SI{36.91}{\%} & 1 \\
 		\hline
	\end{tabular}
\end{table}
\begin{table*}[t!]
\renewcommand{\arraystretch}{1.4}
	\caption{Detection clustering experiments along with the best determined parameters and scores. The \emph{CS.} and \emph{Eqs.} columns list the coordinate systems and the equations utilized for the corresponding experiments. Parameter units are omitted for better legibility. $\epsilon_t=0.25$ for all listed experiments.}
	\label{tab:step1_scores}
	\centering
	\resizebox{.99\linewidth}{!}{%
	\begin{tabular}{r|r|lllll}
		\hline
 		& \textbf{\#} & \textbf{Method} & \textbf{CS.} & \textbf{Eqs.} & $\mathbf{V_1\textbf{-meas.}}$ & \textbf{Optimized Parameter Sets} \\
 		\hline
 		\parbox[t]{2mm}{\multirow{4}{*}{\rotatebox[origin=c]{90}{\textbf{Baseline}}}} & 1 & Expert Setting & CCS & \ref{eq:filter_metric}, \ref{eq:step1_1} & \SI{67.57}{\percent} &  $\epsilon_{xy}=1.00$, $\epsilon_{v_r}=5.00$, $N_{\text{min}}=3$, $v_{r,\text{min}}=0.40$ \\
		& 2 & Baseline & CCS & \ref{eq:filter_metric}, \ref{eq:step1_1} & \SI{68.87}{\percent} &  $\epsilon_{xy}=0.62$, $\epsilon_{v_r}=11.2$, $N_{\text{min}}=3$, $v_{r,\text{min}}=0.23$ \\
		& 3 & Baseline & FCS & \ref{eq:filter_metric}, \ref{eq:step1_1} & \SI{69.19}{\percent} &  $\epsilon_{xy}=0.60$, $\epsilon_{v_r}=12.3$, $N_{\text{min}}=3$, $v_{r,\text{min}}=0.25$ \\
		& 4 & Baseline -- unfiltered & FCS & \ref{eq:step1_1} & \SI{69.19}{\percent} &  $\epsilon_{xy}=0.60$, $\epsilon_{v_r}=12.3$, $N_{\text{min}}=3$, $v_{r,\text{min}}=0.25$ \\
 		\hline
 		\parbox[t]{2mm}{\multirow{4}{*}{\rotatebox[origin=c]{90}{\textbf{Methods}}}} & 5 & Euclidean $xy$-distance & FCS & \ref{eq:filter_metric}, \ref{eq:step1_2} & \SI{69.33}{\percent} &  $\epsilon_{xy}=0.76$, $\epsilon_{v_r}=14.1$, $N_{\text{min}}=3$, $v_{r,\text{min}}=0.31$ \\
 		& 6 & Euclidean $xyv_r$-distance & FCS & \ref{eq:filter_metric}, \ref{eq:step1_3} & \SI{69.47}{\percent} &  $\epsilon_{xyv_r}=0.72$, $\epsilon^{'}_{v_r}=13.5$, $N_{\text{min}}=3$, $v_{r,\text{min}}=0.48$ \\
 		& 7 & Adaptive $N_{\text{min}}(r)$ & FCS & \ref{eq:filter_metric}, \ref{eq:min_pts}, \ref{eq:step1_1} & \SI{70.56}{\percent} &  $\epsilon_{xy}=0.76$, $\epsilon_{v_r}=8.63$, $N_\text{min,50}=3.02$, $v_{r,\text{min}}=0.46$, $\alpha_r=0.99$ \\
  		& 8 & \textbf{Combined $\mathbf{N_{\text{min}}(r)}$ \& $\mathbf{\epsilon_{xyv_r}}$} & \textbf{FCS} & \textbf{\ref{eq:filter_metric}, \ref{eq:min_pts}, \ref{eq:step1_3}} & \textbf{\SI{71.55}{\percent}} &  $\epsilon_{xyv_r}=1.04$, $\epsilon^{'}_{v_r}=1.03$, $N_\text{min,50}=3.87$, $v_{r,\text{min}}=1.00$, $\alpha_r=0.99$ \\ 
  		\hline
  		\parbox[t]{2mm}{\multirow{3}{*}{\rotatebox[origin=c]{90}{\textbf{Control}}}} & 9 & Combined -- unfiltered & FCS & \ref{eq:min_pts}, \ref{eq:step1_3} & \SI{69.91}{\percent} &  $\epsilon_{xyv_r}=1.18$, $\epsilon^{'}_{v_r}=1.01$, $N_\text{min,50}=3.81$, $v_{r,\text{min}}=0.98$, $\alpha_r=0.99$ \\ 
  		& 10 & Best setting -- unfiltered & FCS & \ref{eq:min_pts}, \ref{eq:step1_3} & \SI{69.86}{\percent} &  $\epsilon_{xyv_r}=1.04$, $\epsilon^{'}_{v_r}=1.03$, $N_\text{min,50}=3.87$, $v_{r,\text{min}}=1.00$, $\alpha_r=0.99$ \\ 
  		& 11 & Best setting on baseline & FCS & \ref{eq:filter_metric}, \ref{eq:step1_1} & \SI{67.73}{\percent} &  $\epsilon_{xy}=1.04$, $\epsilon_{v_r}=1.03$, $N_{\text{min}}=4$, $v_{r,\text{min}}=1.00$ \\ 
 	\end{tabular}
	}
\end{table*}
The advantages of detection filtering for clustering parameterization will be discussed in Sec. \ref{subsec:step1_res}.
In terms of time savings, two measures are important:
First, the clustering procedure is compared at a fixed setting with both filtered and unfiltered data over the whole data set.
For over \SI{90}{\percent} of the data the clustering procedure was reduced by \SI{1.8}{\percent} up to \SI{8.3}{\percent} with a total average of \textbf{\SI{3.1}{\percent}} time saving.
Second, based on the best clustering result found in Sec. \ref{subsec:step1_res} the clusters created for the filtered and unfiltered version are counted.
The number of created clusters equals the number of feature vectors that have to be extracted and classified in subsequent processing steps.
It directly impacts the required amount of computations.
Contrary to the small differences in clustering time, the number of clusters is drastically decreased through filtering.
On average, every sequence produces \textbf{\SI{23.6}{\percent}} fewer clusters or \textbf{\SI{28.6}{\percent}} when averaging over all cluster instances, respectively.
This reduces the required computations by approximately one quarter.

\subsection{Point Clustering Results} \label{subsec:step1_res}
During step 1 parameter optimization it is not productive to directly optimize towards a perfect ground truth of the scene.
Instead, the ground truth target data is first pre-clustered using very liberal parameters for an initial clustering ($\epsilon_{xy}\myeq\SI{2}{\meter}$, $\epsilon_{v_r}\myeq\SI{25}{\meter\per\second}$, $\epsilon_t\myeq\SI{0.25}{\second}$, $v_{r,\text{min}}\myeq\SI{0.01}{\meter\per\second}$, $N_{\text{min}}=2$).
The pre-clustering process is executed separately for each ground truth object to ensure that the clusters cannot merge unintentionally.
Setting the pre-cluster parameters to sufficiently high values provides an intermediate data representation that can be well optimized towards without requiring excessive step 1 parameterization.

Early experiments revealed a strong tendency of $\epsilon_t$ values towards values in the range of \SIrange{0.22}{0.27}{\second}.
A fixed duration of \SI{0.25}{\second} was chosen, as the value for $\epsilon_t$ directly impacts the size of the sliding window in which the implemented DBSCAN algorithm operates.
By utilizing a single parameter setting several computational steps require merely a single execution during optimization.

\begin{figure*}[t!]
	\centering	
	\includegraphics[width=.92\linewidth]{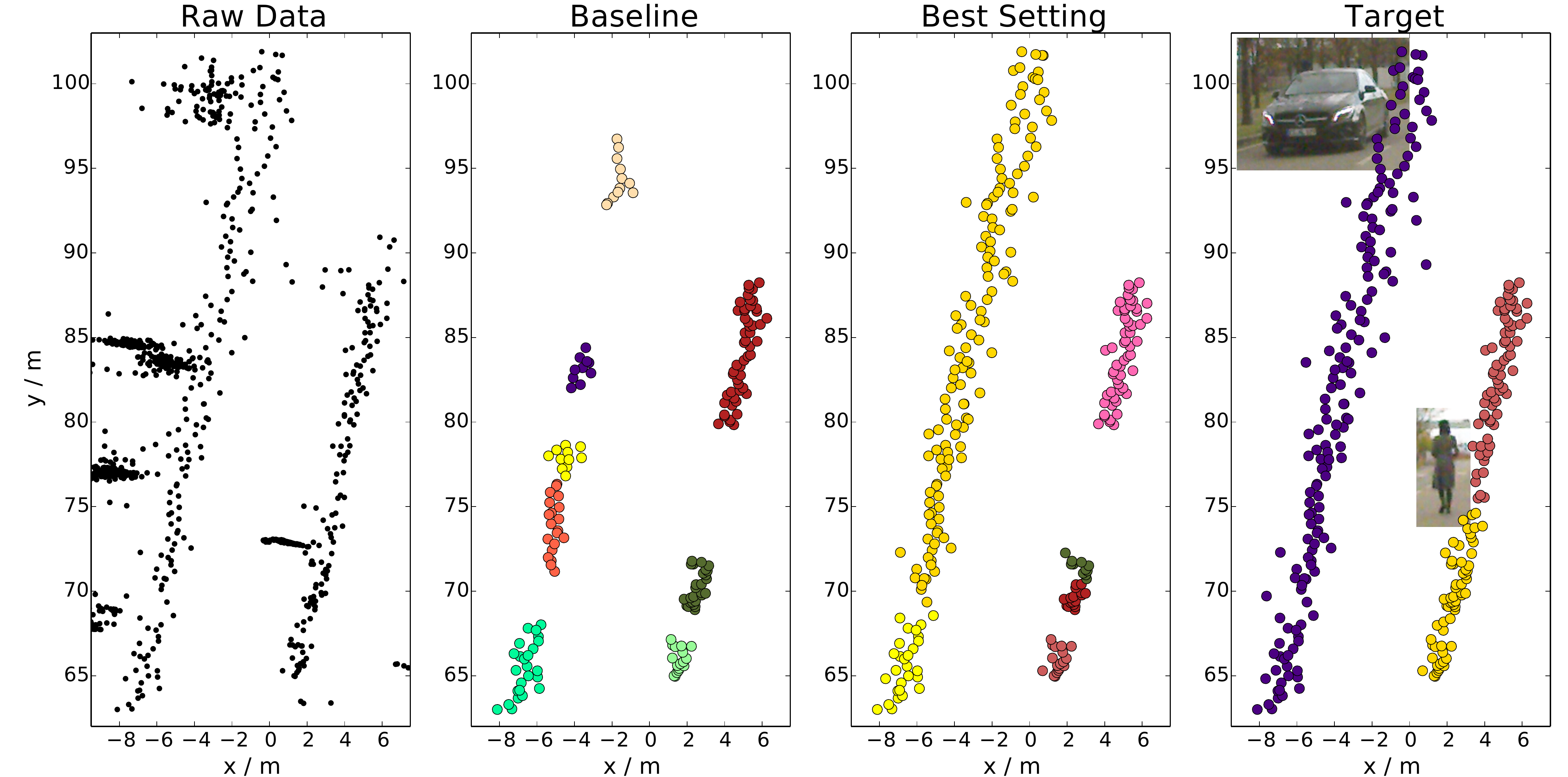}
	\caption{Different processing stages of a \SI{20}{\second} radar measurement sequence with two road user trajectories, a car (left) and a pedestrian (right). Beside raw data and target (ground truth) view, the results from the baseline clustering and the best estimated cluster setting are depicted. Each cluster instances has its own color. Even in the ground truth, the pedestrian instance is labeled as two clusters, because it was not present in the radar data for almost \SI{1}{\second}.}	
	\label{fig:cluster_results}
\end{figure*}

This article comprises several improvements for a standard clustering routine and not all combinations of all proposed methods can be evaluated here.
Tab. \ref{tab:step1_scores} gives an overview of the conducted experiments: \#1 trough \#11.
Initially, four baseline experiments serve as selection basis for fundamental decisions such as the chosen coordinate system or the usage of the filtering method.
Baseline experiments represent a DBSCAN setting with a standard neighborhood criterion (Eq. \ref{eq:step1_1}) and a fixed number of minimum neighbors $N_{\text{min}}$.
As previously found in \cite{Schumann2018b}, an optimized parameter set (\#2) clearly outperforms a fixed expert setting (\#1).
Furthermore, it can be seen that a frame coordinate system is not only the preferred choice in theory but also gives slightly better experimental results (\#3).
When optimizing the baseline setting separately on filtered (\#3) and unfiltered data (\#4) the optimizer selected exactly the same setting, demonstrating the robustness of the method.
Also, the score remains the same for both settings, indicating that very few points of interest are actually removed by filtering.
The next four experiments comprise the three individual adaptations presented in Sec. \ref{subsec:step1}.
The proposed methods are separately added to the filtered baseline method using the FCS.
Then, the best ones among them are combined and evaluated as a whole.
Both alternatives for the distance metric (\#5 and \#6) produce slightly better results than \#3.
Moreover, great improvement can be made by choosing a range-dependent number of minimum neighbors in \#7.
Best results are obtained by combining the range dependency and Euclidean $xyv_r$-distance (\#8) with a $\mathbf{V_1\textbf{-score}}$ of \textbf{\SI{71.55}{\percent}}.
The optimized setting for this last experiment varies greatly from the parameters found in other experiments.
While $\epsilon_{xyv_r}$ is much higher in this case, $\epsilon^{'}_{v_r}$ is remarkably lower.
In order to ensure the validity of these findings, three control experiments are conducted.
The combined approach is optimized again on unfiltered data (\#9).
This time, the results of the unfiltered approach are substantially worse than for the filtered variant, underlining the claim that detection filtering allows selecting wider cluster parameters.
This is backed by the results from another experiment in which the unfiltered data is processed on exactly the setting found to be optimal for the combined approach (\#10).
Note, that this does not contradict the findings for experiments \#3 and \#4.
Apparently, the filtered data can result in the same cluster settings, but is more flexible when used with advanced clustering techniques.
Finally, the $\epsilon$ values are also transferred to the baseline experiment (\#11) under the assumption that this setting is beneficial despite the different meaning of $\epsilon$ parameters in this approach.
In this case as well, however, the setting optimized specifically for this method is better than the setting inferred from \#8.
Another interesting fact about the determined parameters is the tendency of all three experiments with optimized $\alpha_r$ parameters to choose values close to $1$, making the multiplicative factor superfluous. 

In Fig. \ref{fig:cluster_results} the resulting best parameter setting is compared to the baseline parameterization, the ground truth, and the raw data on a short measurement sequence for visual comparison.
For the given scene it can be nicely seen, how better clustering settings help to retain the contours even for remote objects such as the car in the left trajectory.
For the pedestrian represented by the right trajectory the chosen setting can retain more detection points, however one more cluster is formed.
Further, it can be seen that background noise is well suppressed by the filtering and clustering algorithms.

\begin{table*}[!htb]
\renewcommand{\arraystretch}{1.4}
	\caption{Cluster merging experiments along with the best determined parameters and scores. Most parameter units are omitted for better legibility. All experiments are based on the frame coordinate system.}
	\label{tab:step2_scores}
	\centering
	\begin{tabular}{l|l|lll}
		\hline
 		& \textbf{\#} & \textbf{Method} & $\mathbf{V_1\textbf{-meas.}}$ & \textbf{Optimized Parameter Sets} \\
 		\hline
 		\parbox[t]{2mm}{\multirow{2}{*}{\rotatebox[origin=c]{90}{\textbf{Base.}}}} & 3 & Step 1 -- baseline setting & \SI{69.19}{\percent} &  $\epsilon_{xy}=0.60$, $\epsilon_{v_r}=12.3$, $N_{\text{min}}=3$, $v_{r,\text{min}}=0.25$ \\
 		& 8 & Step 1 -- best setting & \SI{71.55}{\percent} &  $\epsilon_{xyv_r}=1.04$, $\epsilon^{'}_{v_r}=1.03$, $N_\text{min,50}=3.87$, $v_{r,\text{min}}=1.00$, $\alpha_r=0.99$ \\
 		\hline
 		\parbox[t]{2mm}{\multirow{2}{*}{\rotatebox[origin=c]{90}{\textbf{Meth.}}}} & 12 & Step 2 -- real velocity estimate & \SI{70.36}{\percent} &  $\epsilon_{d}=1.00$, $\epsilon_{\phi}=\SI{23.11}{\degree}$, $\epsilon_{v}=1.04$, $N_{\text{min}}=1$, $\epsilon_{t2}=0.35$ \\
 		& 13 & \textbf{Step 2 -- cluster continuation} & \textbf{\SI{71.79}{\percent}} & $\epsilon_{\text{pred,min}}=0.94$, $\epsilon_{v}=2.72$, $N_{\text{min}}=1$, $\epsilon_{t2}=0.35$ \\
 	\end{tabular}
\end{table*}

\subsection{Cluster Merging Results}
The evaluation of the second clustering step is closely related to the first one.
Results of both conducted experiments can be found in Tab. \ref{tab:step2_scores}.
For both tested methods the best estimated step 1 settings (\#8) serve as a baseline.
In experiment \#12 the parameter optimization is executed based on the real velocity estimation approach.
In \#13 the cluster center prediction method is tested.
As depicted in Tab. \ref{tab:step2_scores}, the first method actually degrades the results.
Despite finding several beneficial combinations, this approach often also merges undesired clusters, which quickly leads to a score deterioration.
One possible explanation for this is a potential problem with the utilized real velocity estimation algorithm in finding accurate values.
This can be caused, e.g., by a low number of detections in a cluster, or objects such as vulnerable road users which are often detected with multiple different radial velocity values.
In contrast, the monitoring of cluster centers proves to be a more stable solution with fewer merges, but primarily fewer faulty ones.
The total score can be improved slightly from \SI{71.55}{\percent} to \textbf{\SI{71.79}{\percent}}.
A major problem with this method is that the process now depends on how long the base cluster has existed previously.
The primary goal of the first clustering step during two-stage clustering is to identify cluster formations that are most certainly part of the same object.
Therefore, at the second stage, the cluster often just started to exist and no forecast for the future can be made.
Note that it is crucial here to not treat the problem as an ``offline'' problem, i.e., real time requirements forbid the usage of data from later time steps.
Therefore, even though the method is beneficial, it only works in cases where several step 1 clusters have successfully been combined to a cluster of adequate temporal extent.

\section{Conclusion}\label{sec:conclusion}
This article presented a multistage clustering framework that is used to segment automotive radar data.
The assignment of radar detections to clusters is an early step in a traditional classification processing chain.
Hence, good cluster results are essential for minimal error propagation.
The clustering performance is enhanced by first improving the quality of input data with a specialized novel filtering algorithm.
The filter combines spatial and Doppler information to reduce the number of background detections.
Subsequently, a DBSCAN algorithm with specialized distance metric is used to obtain a first clustering of the scene.
An adaptive number of minimum neighbors for cluster formation in DBSCAN provides good results regardless of distance.
When combining all individual components, the overall test results are improved even further and clearly outperform all baseline experiments.
Furthermore, a technique was described which merges the cluster results of the first DBSCAN method to even bigger clusters that better resemble the corresponding road users.
This is done by extracting additional information from those clusters, including the object centers and the real non-radial velocities.
While overall test results could not benefit greatly from the second -- cluster merging -- process, the results  indicate that the two-stage concept works in principle, and that clustering could noticeably benefit from a better real velocity estimation technique.
In our future work, this might be done by using a tracker or high resolution radar sensor with overlapping fields of view.
Also, it is planned to more closely examine how the proposed filtering method can be fine-tuned with distance dependent coefficients such as used for clustering, and the impact of a the proposed clustering framework on a classifier shall be investigated.

\section*{Acknowledgment}
The research for this article has received funding from the European Union under the H2020 ECSEL Programme as part of the DENSE project, contract number 692449.

\bibliographystyle{IEEEtran}
\bibliography{IEEEabrv,mybibfile}

\end{document}